\pdfoutput=1
\documentclass[11pt]{article}
\usepackage[final]{acl}
\usepackage{times}
\usepackage{latexsym}
\usepackage[T1]{fontenc}
\usepackage[utf8]{inputenc}
\usepackage{microtype}
\usepackage{inconsolata}
\usepackage{graphicx}
\usepackage{enumitem}
\usepackage{kotex}
\usepackage[table]{xcolor}
\usepackage{lipsum}

\usepackage{graphicx}   
\usepackage{booktabs}   
\usepackage{multirow}   
\usepackage{float}
\usepackage{tabularx}
\usepackage[table]{xcolor}
\usepackage{placeins}
\definecolor{metacolorab}{HTML}{e6f0fa}
\usepackage{amsmath}
\usepackage{array}

\title{
    \centering
    TORSO: Template-Oriented Reasoning Towards General Tasks
}

\author{Minhyuk Kim, Seungyoon Lee, Heuiseok Lim\thanks{Corresponding author} \\
  Korea University, Republic of Korea \\
  \texttt{\{mhkim0929, dltmddbs100, limhseok\}@korea.ac.kr} \\}

\begin{document}
\maketitle


\begin{abstract}


The approaches that guide Large Language Models~(LLMs) to emulate human reasoning during response generation have emerged as an effective method for enabling them to solve complex problems in a step-by-step manner, thereby achieving superior performance. However, most existing approaches using few-shot prompts to generate responses heavily depend on the provided examples, limiting the utilization of the model’s inherent reasoning capabilities. Moreover, constructing task-specific few-shot prompts is often costly and may lead to inconsistencies across different tasks. In this work, we introduce \textbf{T}emplate-\textbf{O}riented \textbf{R}ea\textbf{so}ning~(TORSO), which elicits the model to utilize internal reasoning abilities to generate proper responses across various tasks without the need for manually crafted few-shot examples. Our experimental results demonstrate that TORSO achieves strong performance on diverse LLMs benchmarks with reasonable rationales.

\end{abstract}

\section{Introduction}

\begin{figure*}[t]
    \centering
    \vspace{1.2em}
    \includegraphics[width=\textwidth]{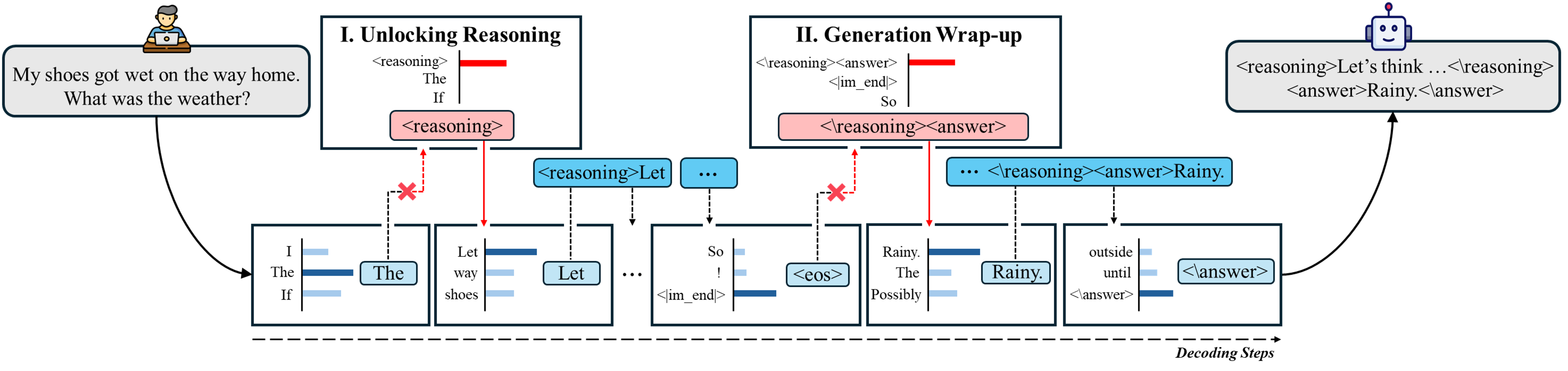}
    \caption{Overview of TORSO.}
    \vspace{1.2em}
    \label{fig:figure_1}
\end{figure*}


As the incorporation of human-like reasoning into Large Language Models~(LLMs) has led to significant performance gains, numerous studies guiding the models to induce reasoning path via in-context learning approaches have been conducted~\cite{chen2023programthoughtspromptingdisentangling,wang2023cuecotchainofthoughtpromptingresponding,wei2022chain}. Most approaches include in-context learning methods such as Chain of Thought~(CoT)~\cite{wei2022chain} and Least-to-Most~(LtM)~\cite{zhou2023leasttomostpromptingenablescomplex}, which guide the model to follow specific reasoning paths through carefully curated few-shot prompts. On the other hand, training methods that leverage curated reasoning datasets based on given queries or instructions~\cite{openai2024reasoning,qwen2025qwq,guo2025deepseek} have become one of the specific approaches to elicit reasoning ability. These methods are fundamentally designed to expand the model’s reasoning process, analogous to how humans approach complex problems step by step~\cite{sun2024surveyreasoningfoundationmodels}.


Such methods have shown their effectiveness in domains that require high-level reasoning, such as mathematics and science~\cite{hwang2024self}. However, since in-context learning relies on imitating the reasoning process presented within the prompt, the design of the prompt plays a critical role in determining the model’s performance~\cite{stechly2024chain}. Due to this dependency, designing prompts that yield consistent performance across various tasks remains a challenging problem~\cite{zhang2025doescotpromptnot}. Furthermore, building reasoning models trained with pre-constructed rationales requires large amounts of additional training data, and they are often targeted at STEM domains, limiting their generalizability to a broader range of tasks~\cite{bae2025exaone}.


In this work, we propose \textbf{T}emplate-\textbf{O}riented \textbf{R}ea\textbf{so}ning~(TORSO), a method designed to guide LLMs to generate responses based on their inherent reasoning ability across various tasks, without relying on additional training data or task-specific few-shot prompts. 
TORSO is founded on the assumption that most LLMs already possess reasoning ability acquired through the vast amounts of training data they process during their learning phase.
TORSO employs logit processing to guide the model's decoding process. We inject a specific token at the initiation stage to unlock reasoning ability and another at the end stage to encourage the model to wrap up the generated rationale into a final answer that directly aligns with the instruction.


Through the experiments on various LLM benchmarks, TORSO consistently outperforms the baselines across the tasks compared to in-context learning methods. Furthermore, we conduct a qualitative evaluation to assess the plausibility of the generated rationales. Our results indicate that the rationales induced by TORSO are appropriate. Our findings suggest that TORSO can effectively guide the models to construct their own reasoning paths without relying on few-shot prompts, and even demonstrate that conventional in-context learning approaches may hinder the coherent reasoning processes.


\section{TORSO}
\begin{table*}[ht]
\centering
\resizebox{1\textwidth}{!}{
\begin{tabular}{cccccccccc}
\toprule
\textbf{Model} & \textbf{Method} & \quad & \textbf{GSM8K} & \textbf{ARC} & \textbf{TruthfulQA} & \textbf{RACE} & \textbf{MMLU} & \textbf{GAOKAO} & \textbf{Avg.} \\
\midrule
\multirow{1}{*}{DeepSeek-R1-Distill-Llama-8B}
  & Base         & \quad & 0.8597 & 0.9226 & 0.4839 & 0.7744 & 0.6935 & 0.4982 & 0.7054\\
\midrule
\multirow{7}{*}{Llama-3.1-8B-Instruct}
  & Base         & \quad & 0.7726 & 0.8493 & 0.4693 & 0.7145 & 0.5788 & 0.4090 & 0.6323 \\
  & CoT-Zero     & \quad & 0.7582 & 0.9125 & 0.4649 & 0.7625 & 0.5987 & 0.4496 & 0.6577 \\
  & CoT          & \quad & 0.7726 & 0.9162 & 0.5760 & 0.7493 & 0.6946 & 0.4369 & 0.6909 \\
  & ToT          & \quad & 0.7953 & 0.9082 & 0.4825 & 0.7570 & 0.6891 & 0.4126 & 0.6741 \\
  & LtM          & \quad & 0.7991 & 0.8923 & 0.5102 & 0.7507 & 0.6874 & 0.4551 & 0.6825 \\
  & \multicolumn{1}{>{\columncolor{metacolorab}}c}{TORSO}
  & \multicolumn{1}{>{\columncolor{metacolorab}}c}{\quad}
  & \multicolumn{1}{>{\columncolor{metacolorab}}c}{\textbf{0.8271}}
  & \multicolumn{1}{>{\columncolor{metacolorab}}c}{\textbf{0.9301}}
  & \multicolumn{1}{>{\columncolor{metacolorab}}c}{\textbf{0.5994}}
  & \multicolumn{1}{>{\columncolor{metacolorab}}c}{\textbf{0.8440}}
  & \multicolumn{1}{>{\columncolor{metacolorab}}c}{\textbf{0.7020}}
  & \multicolumn{1}{>{\columncolor{metacolorab}}c}{\textbf{0.4759}} 
  & \multicolumn{1}{>{\columncolor{metacolorab}}c}{\textbf{0.7298}} \\
\midrule
\multirow{7}{*}{gemma-2-9b-it}
  & Base         & \quad & 0.7665 & 0.9263 & 0.6667 & 0.8572 & 0.6357 & 0.3714 & 0.7040 \\
  & CoT-Zero     & \quad & 0.7688 & 0.9668 & 0.7061 & 0.8684 & 0.7251 & 0.4842 & 0.7532 \\
  & CoT          & \quad & 0.7597 & 0.9482 & 0.7149 & 0.7911 & 0.7412 & 0.5237 & 0.7465 \\
  & ToT          & \quad & 0.7680 & 0.9680 & 0.7339 & 0.8538 & 0.7100 & 0.5200 & 0.7590 \\
  & LtM          & \quad & 0.7695 & 0.9689 & 0.7397 & 0.8809 & 0.7169 & 0.5455 & 0.7702 \\
  & \multicolumn{1}{>{\columncolor{metacolorab}}c}{TORSO}
  & \multicolumn{1}{>{\columncolor{metacolorab}}c}{\quad}
  & \multicolumn{1}{>{\columncolor{metacolorab}}c}{\textbf{0.8188}}
  & \multicolumn{1}{>{\columncolor{metacolorab}}c}{\textbf{0.9705}}
  & \multicolumn{1}{>{\columncolor{metacolorab}}c}{\textbf{0.7427}}
  & \multicolumn{1}{>{\columncolor{metacolorab}}c}{\textbf{0.8893}}
  & \multicolumn{1}{>{\columncolor{metacolorab}}c}{\textbf{0.7473}}
  & \multicolumn{1}{>{\columncolor{metacolorab}}c}{\textbf{0.5474}}
  & \multicolumn{1}{>{\columncolor{metacolorab}}c}{\textbf{0.7860}} \\
\midrule
\multirow{7}{*}{Mistral-7B-Instruct-v0.2}
  & Base         & \quad & 0.3942 & 0.7437 & 0.4254 & 0.6365 & 0.4106 & 0.2652 & 0.4793 \\
  & CoT-Zero     & \quad & 0.4147 & 0.7786 & 0.5351 & 0.6748 & 0.4993 & 0.2828 & 0.5309 \\
  & CoT          & \quad & 0.4193 & 0.7752 & 0.5190 & 0.6957 & 0.5172 & 0.3004 & 0.5378 \\
  & ToT          & \quad & 0.4215 & 0.7790 & 0.4942 & 0.6532 & 0.5127 & 0.3362 & 0.5328 \\
  & LtM          & \quad & 0.4071 & 0.7828 & 0.4401 & 0.6407 & 0.4793 & 0.3186 & 0.5114 \\
  & \multicolumn{1}{>{\columncolor{metacolorab}}c}{TORSO}
  & \multicolumn{1}{>{\columncolor{metacolorab}}c}{\quad}
  & \multicolumn{1}{>{\columncolor{metacolorab}}c}{\textbf{0.4375}}
  & \multicolumn{1}{>{\columncolor{metacolorab}}c}{\textbf{0.7925}}
  & \multicolumn{1}{>{\columncolor{metacolorab}}c}{\textbf{0.5906}}
  & \multicolumn{1}{>{\columncolor{metacolorab}}c}{\textbf{0.7437}}
  & \multicolumn{1}{>{\columncolor{metacolorab}}c}{\textbf{0.5541}}
  & \multicolumn{1}{>{\columncolor{metacolorab}}c}{\textbf{0.3459}}
  & \multicolumn{1}{>{\columncolor{metacolorab}}c}{\textbf{0.5774}} \\
\bottomrule
\end{tabular}
}
\caption{Performance of different methods across various benchmarks, including average score.}
\label{tab:table_1}
\end{table*}


TORSO is based on the assumption that LLMs inherently possess reasoning ability acquired during large-scale pretraining on sufficiently diverse and extensive data, even without explicit reasoning training. Based on this, we focus on eliciting the model's reasoning ability with minimal intervention. To this end, we manipulate the decoding process by forcibly adjusting the probability distribution over token generation. An overview of our pipeline is shown in Figure~\ref{fig:figure_1}.


\paragraph{Step 1: Unlocking Reasoning} We aim to activate the reasoning ability of the model by manipulating token generation probabilities. To be specific, we force the model to generate the token \texttt{<reasoning>} at the first step of decoding in response to a user query. Since LLMs generate outputs in an auto-regressive manner, the presence of \texttt{<reasoning>} at the beginning of the sequence influences the subsequent generation process, encouraging the model to produce reasoning-oriented responses.


As shown in the Unlocking Reasoning part of Figure~\ref{fig:figure_1}, we apply logit processing at the first decoding step of the LLM to forcibly assign very high logits to the tokens composing \texttt{<reasoning>}, regardless of the model’s original probability distribution. This intervention ensures that the output sequence always begins with \texttt{<reasoning>}, providing the model with an explicit reasoning signal.


\paragraph{Step 2: Generation Wrap-up} To ensure that the model produces a final answer that directly corresponds to the user's instruction following a long-form rationale, we inject a token that guides the model to refer its reasoning into a final answer. When the model is about to terminate generation by internally signaling an End-of-Sequence, we insert the \texttt{</reasoning>} token to indicate the end of the rationale, followed by the \texttt{<answer>} token to initiate answer generation.


This encourages the model to interpret the preceding output as a rationale. Consequently, the model incorporates the previously generated content when producing a final, synthesized answer. After generating the \texttt{<answer>} token, the model provides its final response and \texttt{</answer>} to close the answer field. Finally, internal signals end the decoding process by generating End-of-Sequence token.


\paragraph{}
\noindent By inducing the model to make a reasoning process itself, without constructing curated shots or training data for specific tasks, TORSO effectively harnesses the model’s internal ability to form reasonable reasoning.

\section{Experimental Setup}

\subsection{Implementation}


\paragraph{Models} For our experiments, we employ widely used open-source LLMs including Llama-3.1-8B-Instruct~\cite{meta_llama3_1_2024}, Gemma-2-9B-it~\cite{team2024gemma}, and Mistral-7B-Instruct-v0.2~\cite{jiang2023mistral7b}.


\paragraph{Hyperparameters} To ensure consistent experimental conditions for all models, we apply the same decoding hyperparameters across all experiments. We set the maximum generation length to 8192 tokens, the temperature $\tau$ to 1.0, top-$k$ to 50, and top-$p$ to 1.0.
\label{hyperparameters}

\subsection{Evaluation}


\paragraph{Quantitative} To comprehensively evaluate the effectiveness of TORSO, we adopt six benchmark datasets targeting LLMs evaluation. These include GSM8K~\cite{cobbe2021training} and ARC-Easy~\cite{clark2018think} for mathematics and science, TruthfulQA~\cite{lin2022truthfulqameasuringmodelsmimic} for evaluating reliability, RACE~\cite{lai2017race} for reading comprehension, and MMLU~\cite{hendrycks2021measuringmassivemultitasklanguage} and GAOKAO~\cite{zhong2023agieval} for comprehensive knowledge evaluation. We assess model performance based on generated outputs using the Exact Match score as our primary evaluation metric.


\paragraph{Rationale Quality Evaluation} We also conduct a pairwise qualitative evaluation to assess whether the induced reasoning from TORSO is appropriate to the given query. For the judge model, we employ GPT-4o. Samples that both TORSO and the corresponding baselines produce the correct answer are selected from the GSM8K dataset. For each case, the judge evaluates the rationale generated by TORSO and the baseline, determining which output provides a more suitable rationale for the given query.


To ensure a fair comparison with minimal positional bias, we randomly sample 200 questions and conduct 400 comparisons by reversing the order in which rationales are presented in the prompt. Each comparison is repeated five times per sample, and we report the win, tie, and lose ratios accordingly. The detailed prompt used for evaluation can be found in Appendix~\ref{experiments_details_2}.


\subsection{Baselines}
We adopt several major baselines, including the base model performance and representative in-context learning methods such as Chain-of-Thought (CoT)\cite{wei2022chain}, Tree-of-Thought (ToT)\cite{yao2023tree}, and Least-to-Most (LtM)~\cite{zhou2023leasttomostpromptingenablescomplex}, all evaluated under the 5-shot setting. For CoT, we also consider a zero-shot CoT, where the phrase \textit{"Let’s think step by step."} is appended without providing any shots. In our experiments, ToT is applied in a few-shot in-context learning manner~\cite{tree-of-thought-prompting}.

\begin{figure}[t]
    \centering
    \includegraphics[width=0.45\textwidth]{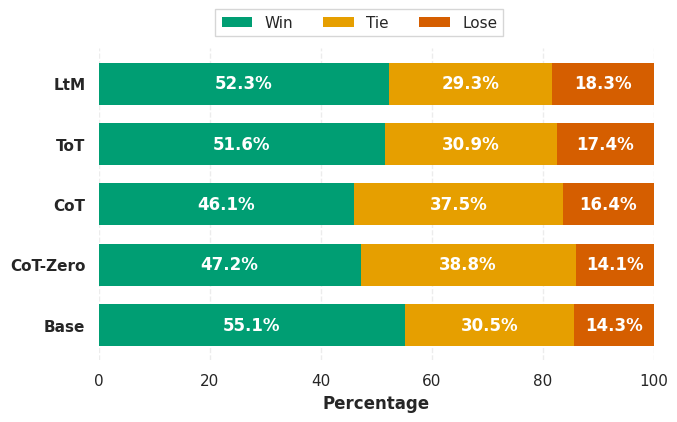}
    \caption{Qualitative comparison results of TORSO against baseline methods on the GSM8K benchmark using Llama-3.1-8B-Instruct. The bars represent the win, tie, and lose percentages.}
    \label{fig:figure_2}
\end{figure}

\section{Results}


\paragraph{Overall Performance} As shown in Table~\ref{tab:table_1}, TORSO consistently exceeds the baselines across the evaluated benchmarks. Specifically, TORSO remains effective even on MMLU and GAOKAO, which include a large number of questions focused on simple factual knowledge such as history and religion. This suggests that the existence of a reasoning can be significant even for questions that are relatively easy in difficulty or have a simple process for generating the correct answer. TORSO achieves these benefits through a minimal intervention, without the need for few-shot prompts or additional data construction.


When comparing CoT-zero and CoT, there are often cases where performance actually decreases when more shots are increased. Notably, for the gemma-2-9b-it model, CoT performs worse than CoT-zero on both GSM8K and ARC in mathematics and science where reasoning is generally known to be effective. This suggests that providing few-shot prompts to help the model reason may actually constrain the model's reasoning path to the shots.


We also observe that a discrepancy in performance can arise between models, depending on the methodology used for shot composition. For instance, with Mistral-7B-Instruct-v0.2, ToT yields better performance than LtM on all benchmarks except ARC, whereas gemma-2-9b-it performs better with LtM. However, TORSO consistently improves performance without relying on any shot composition, as it elicits the model’s reasoning abilities through minimal intervention.

\begin{figure}[t]
    \centering
    \includegraphics[width=0.45\textwidth]{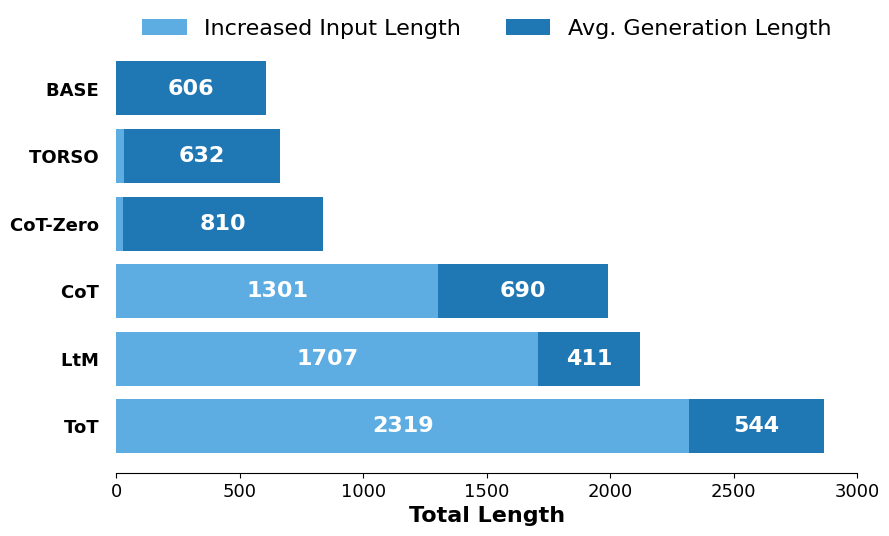}
    \caption{Increased input length and average generation length across methods. Each bar represents the increased input length (light blue) and average generation length (dark blue). Although TORSO does not increase the actual input length, we accounted for the forcibly appended tokens during decoding in our measurement.}
    \label{fig:figure_3}
\end{figure}


\paragraph{Comparison with Reasoning Model} We also compare TORSO with existing reasoning model, DeepSeek-R1-Distill-Llama-8B. On all benchmarks except GSM8K and GAOKAO, TORSO applied to Llama 3.1 8B Instruct achieves superior performance. TORSO demonstrates higher performance even without learning additional reasoning data. This suggests that reasoning models, precisely because they focus on solving problems demanding STEM and complex logical structures, might be vulnerable on general benchmarks. This vulnerability highlights that reasoning models are not universally optimal across general tasks and underscores the value of TORSO.


\paragraph{Rationale Quality Analysis}
TORSO excels not only in its accuracy but also in the quality of the rationales it generates. Given that Figure~\ref{fig:figure_2} presents evaluations restricted to samples with correct answers, TORSO consistently achieves higher win rates across all cases, indicating that it resolves questions with more appropriate rationales. This suggests TORSO helps the model arrive at correct answers and produce reasonable rationales for solving the given tasks.


\paragraph{Length Efficiency} TORSO is cost-efficient when considering both increased input and generation length. Figure~\ref{fig:figure_3} presents a comparison of the rationale lengths generated by each method when producing correct answers. All baselines that rely on few-shot prompting exhibit significant overhead in terms of input length.

\paragraph{}
In terms of generation length, CoT-zero produces the longest rationales, averaging 810.6 in length. Although CoT results in slightly shorter generations, this reduction is primarily due to the inclusion of few-shot prompts. In contrast, TORSO reaches correct answers with rationales that are comparably short to those generated by the base model. LtM and ToT yield generations with lengths of 411 and 544, respectively, which are shorter than TORSO's, but both methods require input lengths exceeding 1500. These results indicate that using a template to guide reasoning offers both high answer accuracy and improved efficiency, especially when compared to methods that rely on few-shot prompts.

\section{Ablation Study}

To better understand how the template used in TORSO affects model performance, we conduct an ablation study using various templates. We evaluate performance under the same hyperparameter settings described in Section \ref{hyperparameters}, using the Llama-3.1-8B-Instruct model across four benchmarks: GSM8K, ARC, TruthfulQA, and RACE. The results are presented in Table \ref{tab:tab2}.

\paragraph{Semantically Similar}
In most configurations that replace the template used in TORSO with semantically similar templates (e.g., \texttt{<think>}, \texttt{<solution>}, \texttt{<result>}), performance surpasses the CoT (5-shot) baseline. These outcomes are consistent with the results observed when using the original \texttt{<reasoning>} + \texttt{<answer>} template in TORSO.

\paragraph{Arbitrary Placeholders}
In the case of arbitrary placeholders, the CoT (5-shot) baseline generally exceeds the performance. While the \texttt{<partⅠ>} + \texttt{<partⅡ>} template achieves a higher score on the RACE benchmark than the CoT (5-shot) baseline, this improvement does not consistently appear across other benchmarks. Although some specific tasks can benefit from arbitrary placeholders, their effectiveness is limited if their semantics do not directly contribute to the model's reasoning process.

\paragraph{Random Tokens}
Our results using templates of random tokens show lower performance than the base model across all benchmarks except TruthfulQA. This suggests that injecting meaningless templates during the decoding may interfere with the generation process.

\paragraph{}
These results indicate that the \texttt{<reasoning>} + \texttt{<answer>} template employed in TORSO consistently improves model performance. Additionally, replacing this template with semantically related alternatives can also lead to performance gains.


\begin{table}[t!]
\centering
\renewcommand{\arraystretch}{1.3}
\newcolumntype{C}[1]{>{\centering\arraybackslash}p{#1}}
\resizebox{\columnwidth}{!}{%
\begin{tabular}{cC{4.3cm}cccc}
\toprule
\textbf{Category} & \textbf{Template} & \textbf{GSM8K} & \textbf{ARC} & \textbf{TQA} & \textbf{RACE} \\
\midrule

Base & -- & 0.773 & 0.849 & 0.469 & 0.715 \\
CoT (5 shot) & -- & 0.773 & 0.916 & 0.576 & 0.695 \\
TORSO (Ours) & \texttt{<reasoning>+<answer>} & \textbf{0.827} & 0.930 & \textbf{0.599} & \textbf{0.844} \\
\midrule
\multirow{4}{*}{Semantically Similar} & \texttt{<think>+<answer>} & 0.820 & \textbf{0.936} & 0.592 & 0.821 \\
 & \texttt{<solution>+<answer>} & 0.814 & 0.911 & 0.585 & 0.826 \\
 & \texttt{<reasoning>+<result>} & 0.826 & 0.927 & 0.599 & 0.831 \\
 & \texttt{<reasoning>+<conclusion>} & 0.796 & 0.932 & 0.598 & 0.827 \\
\midrule
\multirow{2}{*}{Arbitrary Placeholders} & \texttt{<partI>+<partII>} & 0.758 & 0.909 & 0.510 & 0.829 \\
 & \texttt{<marker①>+<marker②>} & 0.749 & 0.840 & 0.518 & 0.742 \\
\midrule
\multirow{2}{*}{Random Tokens} & \texttt{<xyz>+<abc>} & 0.724 & 0.848 & 0.500 & 0.662 \\
 & \texttt{<qwer>+<asdf>} & 0.715 & 0.827 & 0.512 & 0.630 \\
\bottomrule
\end{tabular}%
}
\caption{Performance comparison of different forced token strategies across four benchmarks. TQA refers to TruthfulQA.}
\label{tab:tab2}
\end{table}

\section{Conclusion}


In this paper, we propose TORSO, a method that guides reasoning by enforcing a template during the model's decoding phase, without requiring additional effort such as constructing few-shot prompts or conducting training. Experiments across a diverse set of benchmarks demonstrate that TORSO effectively leverages the model's inherent reasoning ability to solve tasks and exhibits robust applicability across domains.

\section*{Limitations}


While TORSO is an effective method for eliciting the reasoning abilities inherent in language models, it has several limitations. First, its applicability becomes challenging in scenarios that require reasoning over newly emerging information or tasks of extreme difficulty that fall outside the model’s training distribution. This is because TORSO is primarily suited for inducing reasoning grounded in previously learned knowledge and patterns. Furthermore, reasoning models explicitly trained to generate rationales according to fixed templates may not reliably achieve performance gains when exposed to TORSO. 
In summary, TORSO is effective but limited in out-of-distribution reasoning tasks, highlighting the need for further investigation.

\section*{Ethics Statement}
All experiments in this study were conducted with transparency and fairness, using only publicly available datasets intended for academic research. No personally identifiable information (PII) was involved. Furthermore, all models employed in this work are publicly released and accessible.

\section*{Acknowledgements}
This work was supported by the Commercialization Promotion Agency for R\&D Outcomes (COMPA) grant funded by the Korean government (Ministry of Science and ICT) (2710086166). 
This work was supported by Institute for Information \& communications Technology Promotion (IITP) grant funded by the Korean government (MSIT) (RS-2024-00398115, Research on the reliability and coherence of outcomes produced by Generative AI). 
This research was supported by Basic Science Research Program through the National Research Foundation of Korea (NRF), funded by the Ministry of Education (NRF-2021R1A6A1A03045425).
\newpage
\bibliography{anthology}

\appendix

\section{Related Works}


Prior work has explored in-context learning methods that provide human-like reasoning traces as part of the prompt, enabling models to imitate these reasoning processes when generating responses, particularly for solving relatively difficult tasks~\cite{wei2022chain,yao2023tree,jung2022maieuticpromptinglogicallyconsistent}. \citet{wang2024chain} attempt to intervene in the model’s decoding path continuously, aiming to induce CoT reasoning without explicit prompting. However, such approaches do not fully leverage the model’s inherent reasoning ability. Instead, they constrain the model to follow the reasoning path embedded in the prompt and ultimately reducing its behavior to mimicking the provided few-shot prompts~\cite{wei2023symboltuningimprovesincontext}.


On the other hand, there have been efforts to develop specialized reasoning models for solving high-difficulty reasoning tasks by directly injecting rationales into the model through large-scale training data~\cite{zhao2024marcoo1openreasoningmodels,qwen2025qwq,guo2025deepseek,bae2025exaone}. However, reasoning models that aim to maximize reasoning ability through supervised learning are typically targeted toward challenging STEM-related domains, which limits their problem-solving ability to those specific areas.


Inspired by \citet{goyal2024thinkspeaktraininglanguage}, which inserts a token \texttt{<pause>} to induce a delay during both the training and inference stages of LLMs in order to enhance performance through inference time scaling, we propose TORSO, drawing inspiration from the idea that injecting specific tokens at inference can influence the entire decoding process.

\section{Qualitative Evaluation Details}

\subsection{Prompts}
\label{experiments_details_2}

\noindent
\begin{tabularx}{0.45\textwidth}{|X|}

\hline
\\
Choose the better rationale for the given query. Answer with ①, ② or ③ for tie. Print only the answer.\\
Query: [query]\\
① [a]\\
② [b]\\
Answer: 
\newline \\
\hline
\end{tabularx}

\quad
\\
We use special symbols ①,②, and ③ in the outputs generated by GPT-4o to facilitate the extraction of final judgements.

\subsection{Total Results}

\begin{table}[htb]
\centering
\renewcommand{\arraystretch}{1.3}
\resizebox{0.45\textwidth}{!}{
\begin{tabular}{c|c|ccc}
\toprule
\textbf{Model} & \textbf{Comparison} & \textbf{Win} & \textbf{Tie} & \textbf{Lose} \\
\midrule
\multirow{5}{*}{Llama-3.1-8B-Instruct}
& vs. BASE      & 1,103 & 610 & 287 \\
& vs. CoT-Zero  & 944   & 775 & 281 \\
& vs. CoT       & 922   & 751 & 327 \\
& vs. ToT       & 1,033 & 619 & 348 \\
& vs. LtM       & 1,047 & 587 & 366 \\
\midrule
\multirow{5}{*}{gemma-2-9b-it}
& vs. BASE      & 829   & 663 & 508 \\
& vs. CoT-Zero  & 812   & 616 & 572 \\
& vs. CoT       & 714   & 602 & 684 \\
& vs. ToT       & 897   & 521 & 582 \\
& vs. LtM       & 824   & 498 & 678 \\
\midrule
\multirow{5}{*}{Mistral-7B-Instruct-v0.2}
& vs. BASE      & 969   & 595 & 436 \\
& vs. CoT-Zero  & 893   & 553 & 554 \\
& vs. CoT       & 761   & 566 & 673 \\
& vs. ToT       & 913   & 511 & 576 \\
& vs. LtM       & 887   & 478 & 635 \\
\bottomrule
\end{tabular}
}
\caption{Win/Tie/Lose counts from 2,000 pairwise qualitative comparisons (GPT-4o judge) of TORSO against each baseline on GSM8K rationales.}
\label{tab:appen_table_2_grouped}
\end{table}

\onecolumn
\section{Performance across MMLU Subject Categories}
\begin{table*}[htb]
\centering
\resizebox{0.95\textwidth}{!}{
\begin{tabular}{cccccccc}
\toprule
\textbf{Model} & \textbf{Method} & \quad & \textbf{STEM} & \textbf{Other} & \textbf{Social Sciences} & \textbf{Humanities} & \textbf{Avg.} \\
\midrule
\multirow{1}{*}{DeepSeek-R1-Distill-Llama-8B}
  & Base         & \quad & 0.7697 & 0.7074 & 0.7319 & 0.6081 & 0.6935 \\
\midrule
\multirow{7}{*}{Llama-3.1-8B-Instruct}
  & Base         & \quad & 0.5503 & 0.6585 & 0.6324 & 0.5103 & 0.5788 \\
  & CoT-Zero     & \quad & 0.5779 & 0.6176 & 0.6737 & 0.5511 & 0.5987 \\
  & CoT          & \quad & 0.6689 & 0.7557 & 0.7673 & 0.6238 & 0.6946 \\
  & ToT          & \quad & 0.6343 & 0.7547 & 0.7943 & 0.6138 & 0.6891 \\
  & LtM          & \quad & 0.6318 & 0.7499 & \textbf{0.7946} & 0.6132 & 0.6874 \\
  & \multicolumn{1}{>{\columncolor{metacolorab}}c}{TORSO}
  & \multicolumn{1}{>{\columncolor{metacolorab}}c}{\quad}
  & \multicolumn{1}{>{\columncolor{metacolorab}}c}{\textbf{0.6803}}
  & \multicolumn{1}{>{\columncolor{metacolorab}}c}{\textbf{0.7593}}
  & \multicolumn{1}{>{\columncolor{metacolorab}}c}{0.7719}
  & \multicolumn{1}{>{\columncolor{metacolorab}}c}{\textbf{0.6329}}
  & \multicolumn{1}{>{\columncolor{metacolorab}}c}{\textbf{0.7020}} \\
\midrule
\multirow{7}{*}{gemma-2-9b-it}
  & Base         & \quad & 0.6629 & 0.6682 & 0.6991 & 0.5547 & 0.6357 \\
  & CoT-Zero     & \quad & 0.7187 & 0.7612 & 0.8122 & 0.6487 & 0.7251 \\
  & CoT          & \quad & \textbf{0.7371} & 0.7741 & 0.8222 & 0.6693 & 0.7412 \\
  & ToT          & \quad & 0.6977 & 0.7361 & 0.7975 & 0.6438 & 0.7100 \\
  & LtM          & \quad & 0.7035 & 0.7602 & 0.8047 & 0.6400 & 0.7169 \\
  & \multicolumn{1}{>{\columncolor{metacolorab}}c}{TORSO}
  & \multicolumn{1}{>{\columncolor{metacolorab}}c}{\quad}
  & \multicolumn{1}{>{\columncolor{metacolorab}}c}{0.7279}
  & \multicolumn{1}{>{\columncolor{metacolorab}}c}{\textbf{0.7773}}
  & \multicolumn{1}{>{\columncolor{metacolorab}}c}{\textbf{0.8278}}
  & \multicolumn{1}{>{\columncolor{metacolorab}}c}{\textbf{0.6878}}
  & \multicolumn{1}{>{\columncolor{metacolorab}}c}{\textbf{0.7473}} \\
\midrule
\multirow{7}{*}{Mistral-7B-Instruct-v0.2}
  & Base         & \quad & 0.3781 & 0.4026 & 0.4524 & 0.4102 & 0.4106 \\
  & CoT-Zero     & \quad & 0.4738 & 0.5356 & 0.5388 & 0.4665 & 0.4993 \\
  & CoT          & \quad & 0.4554 & 0.5645 & 0.5967 & 0.4752 & 0.5172 \\
  & ToT          & \quad & 0.4932 & 0.5220 & 0.5596 & 0.4891 & 0.5127 \\
  & LtM          & \quad & 0.4164 & 0.5233 & 0.5525 & 0.4444 & 0.4793 \\
  & \multicolumn{1}{>{\columncolor{metacolorab}}c}{TORSO}
  & \multicolumn{1}{>{\columncolor{metacolorab}}c}{\quad}
  & \multicolumn{1}{>{\columncolor{metacolorab}}c}{\textbf{0.4941}}
  & \multicolumn{1}{>{\columncolor{metacolorab}}c}{\textbf{0.6241}}
  & \multicolumn{1}{>{\columncolor{metacolorab}}c}{\textbf{0.6282}}
  & \multicolumn{1}{>{\columncolor{metacolorab}}c}{\textbf{0.4995}}
  & \multicolumn{1}{>{\columncolor{metacolorab}}c}{\textbf{0.5541}} \\
\bottomrule
\end{tabular}
}
\caption{Breakdown of MMLU performance by subject category.}
\label{tab:appen_table_1}
\end{table*}


\section{Qualitative Examples of Reasoning}

\renewcommand{\arraystretch}{1.3}  

\noindent
\begin{tabular}{|>{\centering\arraybackslash}m{0.1\textwidth}|m{0.85\textwidth}|}

\hline
\textbf{Model} & Llama-3.1-8B-Instruct \\
\hline
\textbf{Task} & ARC-Easy \\
\hline
\textbf{Question} & A research scientist writes a paper on the initial regrowth of a forest after a fire has damaged the entire ecosystem. Which title would be best for the paper?

A: Primary Succession

B: Stable Communities

C: Climax Communities

D: Secondary Succession \\
\hline
\textbf{Answer (TORSO)} & <reasoning>

The correct title for the paper would be related to the process of regrowth after a fire has damaged the entire ecosystem. Secondary succession (D) is the process of regrowth and establishment of a new ecosystem in an area that was previously inhabited but has been disturbed or damaged, such as after a fire or clear-cutting. Given the context of the paper, the scientist is studying the regrowth of a forest after a fire, which indicates that the ecosystem was previously inhabited but was damaged. Therefore, the correct title would be related to secondary succession.

</reasoning>

<answer>D</answer> \\
\hline
\textbf{Wrong Answer (CoT)} & A forest fire has completely destroyed the ecosystem. That means all the plants and animals are gone, and the land is now barren.
When life starts to grow again in an area with no organisms or soil, it is called primary succession.
Since the forest is starting from scratch after the fire, this situation fits the definition of primary succession.
Therefore, the best title for the paper would be: A.
\\
\hline
\end{tabular}

\clearpage
\renewcommand{\arraystretch}{1.3}  

\noindent
\begin{tabular}{|>{\centering\arraybackslash}m{0.1\textwidth}|m{0.85\textwidth}|}
\hline
\textbf{Model} & Mistral-7B-Instruct-v0.2 \\
\hline
\textbf{Task} & RACE \\
\hline
\textbf{Question} & For most people, the word "fashion" means "clothes". But people may ask the question, "What clothes are in fashion?" And they use the word "fashionable" in the same way. "She was wearing a fashionable coat. His shirt was really a fashionable colour."

But of course there are fashions in many things, not only in clothes. There are fashions in holidays, in restaurants, in films and books. There are even fashions in school subjects, jobs ... and in languages.

Fashions change as time goes. If you look at pictures of people or things from the past, you will see that fashions have always changed. An English house of 1750 was different from one of 1650. A fashionable man in 1780 looked very different from his grandson in 1860.

Today fashions change very quickly. Some of this is natural. We hear about things much more quickly than in the past. Newspapers, radios, telephones and television send information from one country to another in a few hours.

New fashions mean that people will buy new things, so you see \_ .

By  \_  we can see that fashions change as time goes.

A: finding a job

B: looking at pictures

C: eating in a restaurant

D: wearing a fashionable coat \\
\hline
\textbf{Answer (TORSO)} & <reasoning>

The context of the text suggests that the blank should be filled with an action that can be associated with observing or discovering fashions. Among the given options, "looking at pictures" is the most suitable answer as it aligns with the idea of examining past and present fashions.

The text discusses how fashions change over time and can be observed in various aspects of life, including clothes, holidays, restaurants, films and books, school subjects, jobs, and even languages. The author mentions that people can look at pictures to see how fashions have evolved. Therefore, the correct answer is "looking at pictures."

</reasoning>

<answer>B</answer>\\
\hline
\textbf{Wrong Answer (CoT)} & The passage begins by explaining that many people associate fashion with clothing, and it gives specific examples like a fashionable coat and a fashionable shirt color. Since these examples are mentioned early in the text, it seems like they are important to understanding how fashion works. Wearing a fashionable coat is a direct expression of current fashion trends, so by observing people who wear such clothing, we can understand what is considered fashionable at a certain time. If someone is wearing a coat that matches the popular styles of the time, it reflects the fashion of that era. Therefore, by wearing a fashionable coat, we can see how fashions change as time goes. This makes it a reasonable answer to complete the sentence in the passage.  So the answer is D.
\\
\hline
\end{tabular}

\end{document}